# Estimating the Value of Computation in Flexible Information Refinement

Michael C. Horsch*  
horsch@cs.ubc.ca

David Poole  
poole@cs.ubc.ca

Department of Computer Science  
University of British Columbia  
2366 Main Mall,  
Vancouver, B.C., Canada V6T 1Z4

## Abstract

We outline a method to estimate the value of computation for a flexible algorithm using empirical data. To determine a reasonable trade-off between cost and value, we build an empirical model of the value obtained through computation, and apply this model to estimate the value of computation for quite different problems. In particular, we investigate this trade-off for the problem of constructing policies for decision problems represented as influence diagrams. We show how two features of our anytime algorithm provide reasonable estimates of the value of computation in this domain.

## 1 INTRODUCTION

Anytime algorithms are designed to construct solutions to difficult computational problems by incrementally improving an existing (sub-optimal) solution [Drummond & Bresina, 1990]. This kind of algorithm is interruptible; without interruption, computation may continue well past the point at which the computation is no longer valuable, if one were to consider the cost of computation.

Flexible algorithms are designed to solve difficult computational problems by smoothly trading off the value of the sub-optimal solution with the cost of computing such a solution [Horvitz, 1990; Russell & Wefald, 1992]. The problem faced by flexible algorithms is that the problem of finding an appropriate tradeoff point is a meta-level problem, which can be solved only if computation costs at the meta-level are less than the computation cost at the object level. Usually, simplifying assumptions are made at the meta-level to keep the analysis feasible.

In this paper, we study a particular anytime algorithm for the problem of constructing policies for decision problems represented as influence diagrams [Horsch & Poole, 1998; Horsch, 1998]. This algorithm has a number of general features: the optimal solution is not known before it is computed; the current best solution is incrementally improved, although it is not known in advance how much improvement will be gained by a single computational step; the value of the current best solution is known; the cost of a single computational step is known.

Our trade-off between computational cost and solution value is based on an estimate of the expected value of computation (EVC), which we determine empirically. Using data collected during the course of using information refinement to a large number of simple influence diagrams, we derive a linear model which provides a basis for predicting the expected value of an optimal policy. Based on this prediction, we show how a decision maker can estimate the incremental value of refinement (*i.e.*, the value of doing one more step), making use of information available to the anytime process. The prediction would have negligible cost during the anytime process.

We present some preliminary empirical results using this approach. Our anytime algorithm was applied to number of large influence diagrams which model an agent navigating a maze. For most of these influence diagrams, the optimal policy is not known. The estimated maximum expected value predicted by the model on these influence diagrams is reasonable. The estimate of the incremental value of refinement seems to be somewhat optimistic with respect to value. We are pursuing this issue further.

## 2 BACKGROUND

An influence diagram is a DAG representing a sequential decision problem under uncertainty [Howard & Matheson, 1984]. An ID models the subjective beliefs, preferences, and available actions from the perspective of a single decision maker. A *policy* prescribes an action for each possible combination of observation. An *optimal policy* max-imizes the decision maker's expected value, without regard to the cost of finding such a policy. If computational costs are negligible, the decision maker's expected value depends only on the expected value of an optimal policy. Traditional algorithms which compute the optimal policy using dynamic programming [Howard & Matheson, 1984; Shachter, 1986] usually assume computational costs to be negligible.

## 2.1  FLEXIBLE COMPUTATION

In situations in which there is uncertainty about the state of the world and uncertainty about the possible outcomes of action, it has been argued that a rational decision maker should act so as to maximize expected utility[von Neuman & Morgenstern, 1947; Savage, 1972]. The situation becomes a little more complex when the actions which can be taken include computation.

We treat computation as a meta-level action. That is, the decision maker is faced with a sequential decision problem which has been abstracted in such a way as to ignore computational costs; this level is called the *object level* problem. The decision to invest computational resources towards finding a policy in the object level problem is a meta-level problem. This approach is the basis for *flexible computation* [Horvitz, 1990; Russell & Wefald, 1992].

We also define two kinds of "value" for a policy $\delta$. The first is the *object value* of the policy, $EV_I(\delta)$, which is the expected value of the policy assuming that computational costs on either level are negligible. The second is the *comprehensive value*, $EV_{II}(\delta)$, which includes an accounting for computational costs at the object level.[1]

We consider in this paper those problems for which the comprehensive value is *separable*; that is, the comprehensive value can be separated into two terms, one for the object level value, and one for the computational costs, e.g.:

$$EV_{II}(\delta) = EV_I(\delta) - c(\delta)$$

where $c(\delta)$ is the cost of computing the object level policy. Figure 1 gives a prototypical situation: we show three curves: the object value, the computational costs, and the comprehensive value which is the difference between the object value and computational cost [Horvitz, 1990; Russell & Wefald, 1992].

When computational costs are negligible (*i.e.*, $c(\delta) = 0$), the decision maker maximizes $EV_{II}(\delta)$ by maximizing $EV_I(\delta)$. In Figure 1, this happens at the rightmost edge of the graph. When costs are not negligible, the policy which maximizes $EV_I(\delta)$ may not maximize $EV_{II}(\delta)$, as the cost may be too high. In general, $EV_{II}(\delta) \leq EV_I(\delta)$; *i.e.*, a given policy never increases in value when costs are figured into the value.

## 2.2  INFORMATION REFINEMENT

Information refinement is an iterative approach to constructing policies for decision problems [Horsch & Poole, 1998; Horsch, 1998]. This approach is closely related to the work on compilation of decision models [Heckerman, Breese, & Horvitz, 1989; Lehner & Sadigh, 1993].

The basis of the information refinement algorithm is a process which builds policies in the form of decision trees. A policy is represented by a collection of decision trees, one for each decision node in the influence diagram. These decision trees prescribe actions for contexts which may not make use of all the information available to the decision maker.

The initial policy makes use of none of the information which is available to the decision maker at the time a decision must be made. The available information consists of the agent's observations and previous actions. Each refinement step increases the policy's use of available information; by conditioning action on available information, the process can determine actions which are better suited to more specific situations. The policy is refined by choosing a leaf from one of these trees and applying a single refinement to the leaf, keeping the rest of the policy fixed.

The information refinement algorithm is an anytime algorithm. There is no *a priori* order in which the trees are refined; this is a departure from standard dynamic programming techniques for building an optimal policy. Domain independent heuristics guide the algorithm, applying refinements to decision trees in the problem.

The algorithm always has a current best policy available, which it refines until the decision maker interrupts the process to act. The expected object value of the current best policy is known throughout the anytime process, but the increase in value that may arise in future refinement is not known in advance.

The technique is able to find reasonably good policies for very large problems [Horsch, 1998]. Ignoring computational costs, the value of the policies tends to increase as computational resources are invested in the process. Our approach is able to make decisions with reasonably high expected value with reasonably small computational costs, on problems large enough to make traditional methods infeasible.

---

[1] The subscripts $I$ and $II$ are employed here as a reference to the ideas of Good [Good, 1972], who identified two types of "rationality;" the first, type $I$, is without regard to computational costs, and the second, type $II$, accounting for computational costs.

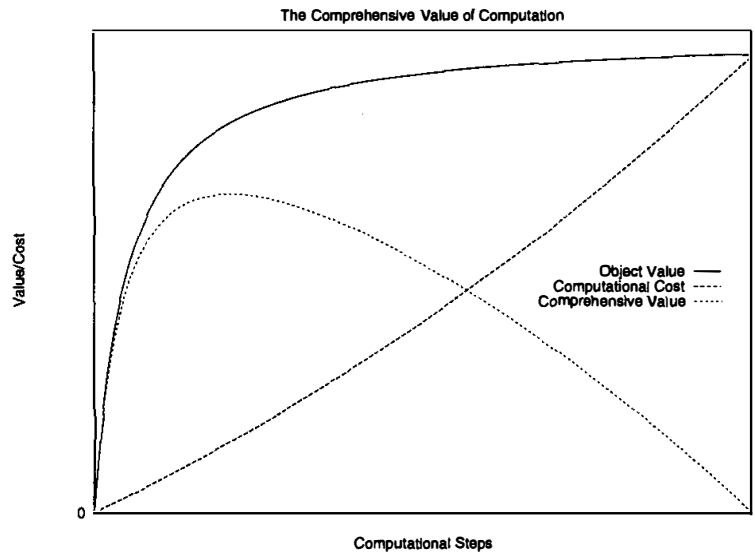

Figure 1: *The comprehensive value of computation for a separable cost function.*

## 3 FLEXIBLE INFORMATION REFINEMENT

In this section we consider the anytime algorithm for information refinement. We show how we have made use of information available during information refinement to estimate the value of computation.

One of the problems faced in our particular situation is that a refinement is not guaranteed to increase the object value of the policy. The value of computation of a single refinement is not necessarily zero, even if it results in no net increase in object value. The investment of computational resources may pay off in future refinement steps.

For example, consider the problem of learning a decision tree representation for the Exor function on two Boolean random variables. Both variables are necessary to represent the Exor function, but individually, neither one provides any information. In the information refinement algorithm, a similar situation arises when no single additional observation increases the value of the current best policy, but observing two (or more) variables would do so.

Thus, the myopic information refinement algorithm is prone to plateaus in which the expected object value does not change as the policy is refined. These plateaus in the object value profile lead to local maxima in the comprehensive value profile if computational costs increase.

Because of the incremental nature of information refinement, we define the *incremental value of computation* for each refinement as the expected object value of the next refinement. This value, $IVC$, depends on knowing the results of future computation. Because there is no randomness in the information refinement process, $IVC$ is determined by the input problem and the information refinement algorithm. Since we do not know $IVC$, we make a simple estimate for it.

At any point in the refinement process, there are a finite number of possible contexts in the current policy refinements which might be refined. Each of these may lead to some increase in object value, although perhaps not immediately. The total object value latent in these possible refinements is $EV_I^* - EV_I$; that is, the optimal expected object value minus the current object value. This value is distributed throughout the possible refinements with some unknown distribution. We make the simple assumption that the total latent object value is distributed uniformly over the possible refinements. The latent value in any single refinement step is

$$LVR = (EV_I^* - EV_I)/n$$

where $n$ is the number of contexts which can be refined in the current policy. We use $LVR$ as an estimate for $IVC$.

It may seem that we have traded one unknown quantity, $IVC$, for another, $EV_I^*$ ($EV_I$ is known). In the next section, we will estimate $EV_I^*$ based on data gathered by applying information refinement to single stage influence diagrams.

## 4 EMPIRICAL RESULTS

We applied information refinement to one hundred randomly constructed single decision influence diagrams. The data we collected was used to learn a linear model to predict an estimate for expected value. This model was applied to much larger influence diagrams. We describe our experiment in detail below.

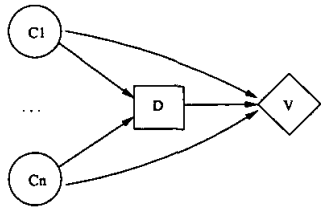

Figure 2: This template influence diagram has one decision node and $n$ informational predecessors.

## 4.1 SAMPLE INFLUENCE DIAGRAMS

The influence diagrams are randomly sampled from a class of diagrams with very specific properties, which we discuss here. A template problem for this class of influence diagram is pictured in Figure 2. For brevity, this class is called the "1-ID($n$)" class, where $n$ is the number of chance nodes.

We use this class of influence diagram because the sample space of all influence diagrams is very large. We also wish to avoid creating essentially random problems with no properties in common with "real" problems. Many of our choices for sampling from this class are based on simplicity, all other things being equal.

The 1-ID($n$) class has the property that all the chance nodes are parents of the decision node and the value node. As well, chance nodes in the 1-ID($n$) class are conditionally independent.

We point out that any influence diagram with a single decision node can be reduced to one in which the only chance nodes are information predecessors (by summing out all the chance nodes which are not information predecessors (using variable elimination, for example [Zhang & Poole, 1996]).

The conditional independence between information predecessors is used to keep the sample space as simple as possible. As well, we consider only binary-valued chance nodes, and a binary-valued decision node.

The 1-ID($n$) class permits some interesting variation in terms of the probability distributions for the chance nodes. For this experiment, the prior probability distribution for each chance node was selected at random from a uniform distribution: for each chance node $C_i$ in the influence diagram, one parameter $x_i$ was drawn from $[0, 1]$ with a uniform distribution. The conditional probability table for the chance node given to the chance node is $(x_i, 1 - x_i)$.

The 1-ID($n$) class also permits some variation in terms of the dependency of the value function on its inputs, i.e., the chance nodes plus the decision node. In these influence diagrams, a value function has $n + 1$ inputs, but may not depend functionally on all of these inputs. In particular, there may be combinations of a subset of the inputs which render the remainder irrelevant. For example, a decision maker's preference may depend on chance node $A$ when $B$ is true, but may not depend on $A$ when $B$ is false.

In order to construct value functions with varying dependencies on its inputs, the following procedure was used. The value function is constructed as a tree, with the inputs as internal nodes, and real values as leaves. The parent nodes of the value node were represented in a list. With probability $b$, the first of these nodes would be used to split the value tree at the current position; with probability $1 - b$, the first node was discarded. This procedure was repeated for every node in the list. The decision node was always used as the last split (i.e., with probability 1), meaning that the actions of the agent always always had an effect on the value. The leaves of the value tree were selected from $[0, 1]$ with a uniform probability distribution.

By varying the parameter $b$, value functions with more or less dependence on its inputs can be constructed. In the experiment described here, the value $b = 0.7794$ was used. This results in value functions (when represented as trees) which are expected to have 200 internal nodes. When $n = 8$, as in our experiment, a value function could have as many as 511 internal nodes. Thus, the value functions used in this part of the experiment are expected to have a significant degree of structure. The information refinement process exploits this kind of structure.

## 4.2 THE TRAINING DATA

One hundred influence diagrams with the properties described above were constructed, and information refinement was applied to each. The process uses heuristic information as guidance; in the experiment described here, we used a heuristic which we call "the second best action heuristic," which has been described in previous work [Horsch & Poole, 1998]. It is used to determine which part of the policy to refine, and is based on the observation that if there is a large difference in expected value between the best action and the second best action in any context, it is probably the case that a refinement to the context will not lead to significant improvements to the value of the policy. The heuristic value $H$ for a context is computed as follows:

$$H = p \frac{v}{v^*}$$

where $p$ is the marginal probability of the given context, $v^*$ is the expected value of the best action in the given context, and $v$ is the expected value of the second best action in the context. We note that $H$ will tend to decrease: as contexts get more specific, $p$ will decrease. As well, $H$ need not converge to zero, as there need not be an action that results in an expected value of $v = 0$.

The following quantities were recorded at each step of the information refinement process: the object value of the current best policy, $EV_I$; the heuristic value for the current refinement step $H$; the number of possible myopic refinements which can be made at the current step. The experiment also determined the optimal expected value, $EV_I^*$ for each influence diagram.

The data collected for each influence diagram in the sample set contains a profile for each step in the refinement process. The points in the profile are not independent in a probabilistic sense. One data point was extracted from each profile, so that the data would be independent. We chose to extract the point in the profile after the tenth refinement step. The profiles at the tenth step have not yet converged; we want to avoid training on data from the regions of the profile at which the process has converged. The tenth step was chosen because the process converges to the optimal policy for these problems after about 60 refinement steps on average.

Three linear models were fit to the data, and a least squares estimate was made for the parameters of polynomial surfaces of degree 1, 2 and 3. The dependent variable was the maximum expected object value, the quantity we wish to predict; the independent variables were the heuristic measure, $H$, which guides the refinement process, and the object value of the current policy $EV_I$.

The three models were examined informally for evidence of over-fitting, and the surfaces of degree 2 and 3 were rejected by geometric considerations. While the sum of squares error for these surfaces was quite small for the training data, the surfaces did not make reasonable extrapolations outside the range of the data. The high degree surfaces extended into negative values, and positive values greater than 1.

The remaining model, a plane in 3-space, had the following form:

$$EV_I^* = c_0 + c_1 EV_I + c_2 H$$

where

$$c_0 = 0.1328$$
$$c_1 = 0.8415$$
$$c_2 = 0.1753$$

Observe that $EV_I$, the object value of the current best policy, is the biggest factor in the prediction of the value of the optimal policy. This agrees with our intuitions: at the start of the refinement process, the current policy is relatively far from optimal, and the heuristic value should be high. As the refinement process proceeds, the current policy converges to optimal, and the heuristic value decreases.

The sum of squares error was quite small: 0.0634, over 100 data points. We applied this model to all the data in all the profiles collected from the single stage influence diagrams. The sum of squares error for the 25300 data points from all the profiles was very small as well: 29.4.

### 4.3 TESTING THE MODEL

The simple linear model obtained in the previous section was applied to 16 multi-stage influence diagrams. Each of these influence diagrams model an agent navigating a maze; we modelled 4 different agents, which vary in the noisiness of actuators and sensors, and four different mazes, which vary in topology. The decision problem is to determine a policy which gets the agent to a goal location in the maze, starting from anywhere in the maze.

Each is a ten stage influence diagram, which implies that the agent must arrive at the goal within ten steps to achieve the reward of 1 (the maximum value); being in any other location is worth nothing to the agent. The information available to the agent is in the form of 4 sensors, one for each compass direction. The agent cannot directly observe its position.

The information space of these problems contains about $2^{60}$ states. For two of the 16 problems, an optimal policy is known; in both problems, the agent has noiseless sensors and actuators, and a policy was constructed which guarantees the agent will arrive at the goal position from any other position in the maze. However, for the remaining 14 problems, an optimal policy is not known. These problems are described in more detail in [Horsch, 1998].

The information refinement procedure was applied for 30 refinement steps on each influence diagram. The average time required for these steps was 20.6 minutes.

The linear model (determined in the previous section based on the 1-ID($n$) data) predicted optimal policy values which were on average 0.19 higher than the current policy at each refinement step; (std. dev. 0.047). We emphasize that this is not an error measurement; the optimal policy may be higher than any policy we have constructed. The average difference between the estimated value of the optimal policy, and the best known policy for these problems is 0.027 (std. dev. 0.14); the estimate is often low at the beginning of the refinement process, and increases with time.

On the two influence diagrams whose optimal policy is known to be 1.0, the initial estimates of the optimal expected value were 0.334 and 0.327; after 30 refinement steps, the estimates were 1.07 and 1.08, respectively. Table 4.3 summarizes the results for all 16 influence diagrams (see the rows labelled (1-ID($n$)).

The difference between the estimates and the known values may be due to the fact that the optimal policy is unknown (and the best policies found are about this far from optimal). On the other hand, the estimates may be inaccurate with respect to these problems because the training data is not a good model for the larger influence diagrams.

To investigate these possibilities further, we constructed 40 new influence diagrams similar to the 16 test problems.

|  |  | Agent 1 | Agent 2 | Agent 3 | Agent 4 |
|---|---|---|---|---|---|
|  | 1-ID($n$): | 1.07 | 0.874 | 0.971 | 0.801 |
| Maze 1 | Similar: | 1.04 | 0.824 | 0.931 | 0.745 |
|  | Best Known: | 1.0 | 0.767 | 0.883 | 0.685 |
|  | 1-ID($n$): | 0.677 | 0.629 | 0.691 | 0.559 |
| Maze 2 | Similar: | 0.599 | 0.549 | 0.617 | 0.475 |
|  | Best Known: | 0.500 | 0.453 | 0.530 | 0.381 |
|  | 1-ID($n$): | 0.879 | 0.726 | 0.785 | 0.625 |
| Maze 3 | Similar: | 0.821 | 0.660 | 0.719 | 0.545 |
|  | Best Known: | 0.741 | 0.584 | 0.635 | 0.450 |
|  | 1-ID($n$): | 1.08 | 0.933 | 0.951 | 0.821 |
| Maze 4 | Similar: | 1.05 | 0.885 | 0.903 | 0.761 |
|  | Best Known: | 1.00 | 0.822 | 0.838 | 0.686 |

Table 1: A summary of the estimated expected value of the optimal policy for the test set of 16 large influence diagrams. The estimates were based on the 1-ID($n$) data, and also on problems similar to the test set. The final estimates are shown, and the value of the best known policy is provided for comparison.

The new problems were smaller instances of the test set (a smaller maze size, and only 5 stages). Information refinement was applied to these smaller influence diagrams and data were collected as for the 1-ID($n$) problems. We fit a linear model to the data, and the model had the following form:

$$EV_I^* = c_0 + c_1 EV_I + c_2 H$$

where

$$c_0 = 0.0388$$
$$c_1 = 0.9252$$
$$c_2 = 0.1384$$

We applied this model to the 16 larger influence diagrams, as before. Table 4.3 summarizes the results (see the rows labelled Similar). In general, the estimates of $EV_I^*$ are smaller using problems similar to the test set than the 1-ID($n$) problems.

### 4.4 USING THE MODEL

We were interested in using our model to estimate the comprehensive value of computation. We used the estimate of the object value of the optimal policy at each refinement step to determine the incremental cost of computation, as outlined above. Figure 3 shows a typical result. In this plot, we plot value as a function of the number of refinement steps in our process. The object value of the current best policy is increasing. We also have provided a cost model for this example, that increases exponentially with the number of refinement steps. The comprehensive value of the policy consists of the difference between the object value and its cost at each step. We note that the object value and the comprehensive value profiles are retrospective; a decision maker faced with a resource bounded problem will not see the entire profile, but only that part which it has computed during information refinement. The comprehensive value is maximized at 3 refinement steps for this particular problem and the given cost function.

The figure also shows the two estimates of the value of the optimal policy. Note how the estimates are higher than the current object value throughout the profile. As indicated above, the object value of the optimal policy is not known for this problem. After 30 refinement steps, the 1-ID($n$) data predicts an optimal value of roughly 0.73; the data based on the smaller mazer walking problems predicts an optimal value of about 0.66. Neither of these estimates are unreasonable for this problem. The best known policy (after 40 refinement steps) is 0.681.

Figure 4 shows the latent value of refinement as predicted by the two data sets (Section 3. Latent value is computed using the estimate of the value of the optimal policy, as described in Section 3. This quantity is decreasing, and is an estimate intended to model the value of future refinements made possible by the current refinement step. Note the small scale of the values, which reflects the fact that the difference between the current object value and the estimated optimal value is assumed to be distributed evenly across the possible refinements.

The graph also shows the differential value for each refinement step. This value represents the difference between the latent value of refinement and the cost of performing a single refinement step. When this difference is positive, the comprehensive value of the next step is expected to increase; when it is negative, the comprehensive value is expected to decrease.

As the graph shows, our estimate of $LVR$ does not predict the maximum comprehensive value for this problem. At 10 refinement steps, the differential value predicted by the 1-ID($n$) data set goes negative. This comes 7 steps after the global maximum in comprehensive value attained. The differential value based on the smaller maze walking problems predicts a zero after 6 refinement steps. In this example, our models over-estimates the incremental value of refinement, and therefore the differential value reaches zero after the comprehensive value is maximized. We are investigating this issue further.

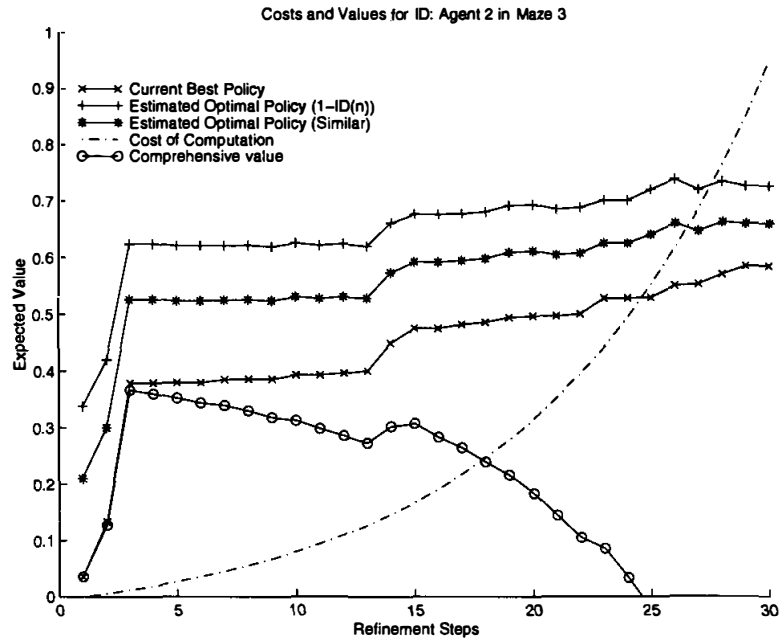

Figure 3: *Value functions for one of the 16 large influence diagrams. Also shown are the object value of the current best policy, and the estimated optimal value functions based on the two data sets. A prototypical cost function is given, and the comprehensive value of the current best policy is derived using this cost function.*

## 5 CONCLUSIONS AND FUTURE WORK

In this paper we have looked at the problem of using available information to estimate when to interrupt an anytime algorithm. Our approach estimated the expected value of the optimal policy from empirical data, and from this, derived an estimate of the value of an investment of computational resources.

Our investigation is specific to the information refinement process. We derived two models for the dependence of expected value of an optimal policy, which is not known during the information refinement process, on measures which are known during the process. The first model was based on data collected while applying information refinement on a large number of simple influence diagrams which could be solved optimally.

This model was applied to a test set of influence diagrams for which finding an optimal policy is infeasible. The estimated value of the optimal policies for the larger influence diagrams was consistently higher than the value for the best known policy during information refinement. However, the estimated values were not unreasonably high.

A second model was based on problems which were similar to the test set; these were smaller than the test set, but still too large to solve for the optimal policy. Again, the estimated value of the optimal policy was consistently higher than the best known policy, but not unreasonable.

We believe that these preliminary results are encouraging. We have shown that a reasonably predictive model can be derived from information which is available to the decision maker during the information refinement process. This data can be collected by a decision making agent, and used to improve future comprehensive performance. More sophisticated learning techniques could be used to provide more accurate estimates.

### Acknowledgments

The authors would like to thank Brent Boerlage of Norsys Software Corp. for advice and support in the use of the Netica API as our Bayesian network engine. The second author is supported by the Institute for Robotics and Intelligent Systems, Project IC-7, and the National Sciences and Engineering Council of Canada Operating Grant OGPOO 44121.

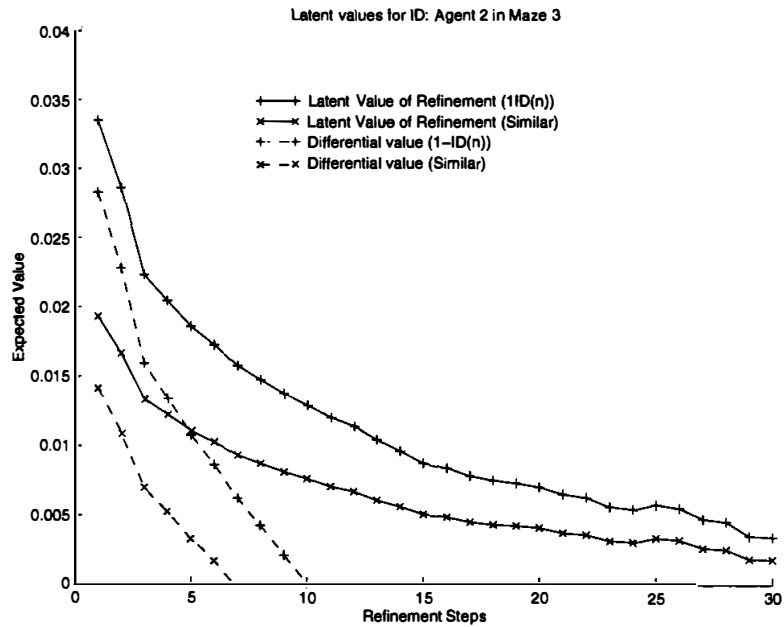

Figure 4: *The latent value of refinement for one of the test problems, derived from the estimates made by the two data sets.*

## References

[Drummond & Bresina, 1990] Drummond, M., and Bresina, J. 1990. Anytime synthetic projection: Maximizing the probability of goal satisfaction. In *Proceedings of the Eighth National Conference on Artificial Intelligence*, 138–144.

[Good, 1972] Good, I. J. 1972. Twenty-seven principles of rationality. In Godambe, V. P., and Sprott, D., eds., *Foundations of Statistical Inference*. Toronto: Holt,Rinehart,Winston. 108–141.

[Heckerman, Breese, & Horvitz, 1989] Heckerman, D. E.; Breese, J. S.; and Horvitz, E. J. 1989. The compilation of decision models. In *Uncertainty in Artificial Intelligence* 5, 162–173.

[Horsch & Poole, 1998] Horsch, M. C., and Poole, D. 1998. An anytime algorithm for decision making under uncertainty. In *Proceedings of the Fourteenth Conference on Uncertainty in Artificial Intelligence*, 246–255.

[Horsch, 1998] Horsch, M. C. 1998. *Flexible Policy Construction by Information Refinement*. Ph.D. Dissertation, Department of Computer Science, University of British Columbia.

[Horvitz, 1990] Horvitz, E. J. 1990. Computation and action under bounded resources. Technical Report KSL-90-76, Departments of Computer Science and Medicine, Stanford University.

[Howard & Matheson, 1984] Howard, R., and Matheson, J., eds. 1984. *Readings on the Principles and Applications of Decision Analysis*. CA: Strategic Decisions Group.

[Lehner & Sadigh, 1993] Lehner, P. E., and Sadigh, A. 1993. Two procedures for compiling influence diagrams. In *Proceedings of the Ninth Conference on Uncertainty in Artificial Intelligence*, 335–341.

[Russell & Wefald, 1992] Russell, S., and Wefald, E. 1992. *Do the Right Thing: Studies in Limited Rationality*. Cambridge, Mass.: MIT Press.

[Savage, 1972] Savage, L. J. 1972. *The Foundations of Statistics*. Dover Publications, Inc.

[Shachter, 1986] Shachter, R. D. 1986. Evaluating influence diagrams. *Operations Research* 34(6):871–882.

[von Neuman & Morgenstern, 1947] von Neuman, J., and Morgenstern, O. 1947. *The Theory of Games and Economic Behaviour*. Princeton University Press.

[Zhang & Poole, 1996] Zhang, N. L., and Poole, D. 1996. Exploiting Causal Independence in Bayesian Network Inference. *Journal of Artificial Intelligence Research* 5:301–328.